\def\year{2022}\relax
\documentclass[letterpaper]{article} 
\usepackage{aaai22}  
\usepackage{times}  
\usepackage{helvet}  
\usepackage{courier}  
\usepackage[hyphens]{url}  
\usepackage{graphicx} 
\usepackage{natbib}  
\usepackage{caption} 
\DeclareCaptionStyle{ruled}{labelfont=normalfont,labelsep=colon,strut=off} 
\usepackage{algorithm}
\usepackage[noend]{algpseudocode}
\algrenewcommand\algorithmicensure{\textbf{Output:}}
\usepackage{newfloat}
\usepackage{listings}
\floatstyle{ruled}
\newfloat{listing}{tb}{lst}{}
\floatname{listing}{Listing}
\nocopyright
\usepackage{amssymb}
\usepackage{multirow}
\usepackage{mathtools}
\usepackage{pifont}
\usepackage{tikz}
\usetikzlibrary{shapes,decorations,arrows,calc,arrows.meta,fit,positioning}
\newcommand{\cmark}{\ding{51}}%
\newcommand{\xmark}{\ding{55}}%
\usepackage{pgfplots}

\usepackage{caption}
\usepackage{subcaption}

\usepackage{xcolor}

\newcommand{\smp}[0]{\textsc{sm}ProbLog}
\newtheorem{example}{Example}

\newtheorem{definition}{Definition}

\title{\smp{}: Stable Model Semantics in ProbLog and its Applications in Argumentation}
\author{
    Pietro Totis,
    Angelika Kimmig,
    Luc De Raedt
}
\affiliations{
    KU Leuven, Dept. of Computer Science; Leuven.AI, B-3000 Leuven, Belgium\\
    firstname.lastname@cs.kuleuven.be
}

\begin{document}

\maketitle

\begin{abstract}
    We introduce \smp{}, a generalization of the probabilistic logic programming language ProbLog. A ProbLog program defines a distribution over logic programs by specifying for each clause the probability that it belongs to a randomly sampled program, and these probabilities are mutually independent. The semantics of ProbLog is given by the success probability of a query, which corresponds to the probability that the query succeeds in a randomly sampled program. 
    It is well-defined when each random sample uniquely determines the truth values of all logical atoms. Argumentation problems, however, represent an interesting practical application where this is not always the case. \smp{} generalizes the semantics of ProbLog to the setting where multiple truth assignments are possible for a randomly sampled program, and implements the corresponding algorithms for both inference and learning tasks. We then show how this novel framework can be used to reason about probabilistic argumentation problems.
    Therefore, the key contribution of this paper are: a more general semantics for ProbLog programs, its implementation into a probabilistic programming framework for both inference and parameter learning, and a novel approach to probabilistic argumentation problems based on such framework.
\end{abstract}

\section{Introduction}
\begin{table*}[h]
    \centering
     \begin{tabular}{| c | c | c | c | c | c |} 
     \hline
     Framework & Epistemic & PLP & Bayesian  & EM Learning & Implementation \\ \hline
     \cite{DBLP:conf/tafa/LiON11} &  \xmark & \xmark & \xmark & \xmark & \cmark \\\hline
     ~\cite{DBLP:conf/ijcai/KidoO17} & \xmark & \xmark & \cmark & \xmark & \xmark \\ \hline
     \cite{DBLP:journals/ijar/MantadelisB20} & \xmark & \cmark & \cmark & \xmark & \cmark \\\hline
     \cite{DBLP:journals/ai/HunterPT20} & \cmark & \xmark & \xmark & \xmark & \xmark \\\hline
     \smp{} & \cmark & \cmark & \cmark & \cmark & \cmark \\
     \hline
     \end{tabular}
     \caption{Argumentation frameworks comparison.}
     \label{tab:compare}
\end{table*}
Designing systems that are able to argue and persuade is a very relevant and challenging task in the development of artificial intelligence. 
The goal of abstract argumentation~\cite{DBLP:journals/ai/Dung95} is to compute the set of acceptable arguments from a framework describing an argumentation process. 
In real-world scenarios the information available is often incomplete or questionable, therefore the ability to take into account uncertainty is fundamental in such situations. 
Consider for instance the following example of argumentative microtext adapted from~\cite{DBLP:conf/lrec/StedeAPAP16}: 
\begin{example}\label{ex:microtext}
Yes, it's annoying and cumbersome to separate your rubbish properly all the time ($a_1$), but small gestures become natural by daily repetition ($a_2$). Three different bin bags stink away in the kitchen and have to be sorted into different wheelie bins ($a_3$). But still Germany produces way too much rubbish ($a_4$) and too many resources are lost when what actually should be separated and recycled is burnt ($a_5$). We Berliners should take the chance and become pioneers in waste separation! ($a_6$)
\end{example} 
In the epistemic approach~\cite{DBLP:journals/ijar/Hunter13} a probabilistic argument graph describes degrees of belief associated to arguments $a_i$, e.g. ``$a_1$ is believed with probability $0.7$'' and their relations, for example $a_1$ attacks $a_2$,  $a_3$ supports $a_1$\dots We want to reason over these elements in order to answer queries like "What is the likelihood of accepting argument $a_6$?". This is a typical probabilistic logic programming (PLP) task: PLP frameworks and languages are designed to provide powerful general-purpose tools for modeling of and reasoning about structured, uncertain domains, e.g. PRISM~\cite{DBLP:conf/ilp/SatoK08}, ICL~\cite{DBLP:conf/ilp/Poole08}, ProbLog~\cite{DBLP:conf/ijcai/RaedtKT07} or LPAD/CP-logic~\cite{DBLP:journals/tplp/VennekensDB09}. 
When considering these two domains, the question is natural: can PLP effectively model argumentation processes and reason over its intrinsic uncertainty? 

This question has been only partially investigated so far: of the two main interpretations of probabilities in argument graphs~\cite{DBLP:journals/ijar/Hunter13}, namely the \emph{constellations} approach and the \emph{epistemic} approach, only the former has been studied in the context of PLP~\cite{DBLP:journals/ijar/MantadelisB20}. In fact, probabilistic argumentation systems propose different combinations of argumentation frameworks, probability interpretations and reasoning systems, tailored to manipulating probabilities in argumentation. As we summarize in Table~\ref{tab:compare}, some do not follow the epistemic approach~\cite{DBLP:conf/tafa/LiON11,DBLP:conf/ijcai/KidoO17,DBLP:journals/ijar/MantadelisB20}, others do not consider PLP or a Bayesian view (i.e. causal influence between random variables defining a single probability distribution) of probabilities~\cite{DBLP:journals/ai/HunterPT20}.
This paper fills this gap and shows how PLP can be used to reason about an epistemic argumentation graph. Approaching argumentation from PLP raises also the question: can we answer queries of the kind ``How does my belief in $a_1$ change if $a_5$ is accepted?'' or perform typical Bayesian tasks as learning the beliefs of an agent given a set of observations of accepted arguments?

We propose a new PLP framework where this is possible, since we argue that existing systems are limited in what problems they can model and reason about. 
In fact, traditional PLP frameworks do not consider as valid inputs the programs modelling cyclic relations involving negations. This is a pattern often found in argument graphs, where reciprocal attacks are common. For example, we will encode this pattern, i.e. accepting $a_1$ inhibits my belief in $a_2$ and vice versa, with the logic rules $\lnot a_1\leftarrow a_2.\; \lnot a_2 \leftarrow a_1$. A model containing such rules would not be a valid input for traditional PLP frameworks.
For this reason, in sections~\ref{sec:semantics} and~\ref{sec:implementation} we propose a PLP system, \smp{}, based on a new semantics, where it is possible to reason over such models. In Section~\ref{sec:args} we show how this framework can model and reason over probabilistic argumentation problems, and how to apply typical PLP techniques in this setting.

The key contributions of the paper are: 
1) we define a novel semantics for PLP based on stable model semantics, 
2) we develop an implementation of a PLP framework, \smp{}\footnote{https://github.com/PietroTotis/smProblog}, derived from ProbLog2~\cite{DBLP:conf/pkdd/DriesKMRBVR15} where inference and learning tasks can be performed under the new semantics, 
3) we show an application of \smp{} to encode and solve probabilistic argumentation problems in a novel reasoning framework based on a Bayesian interpretation and manipulation of probabilities. 

\section{Related work}

In this paper we present an extension of the ProbLog system and semantics~\cite{DBLP:conf/ijcai/RaedtKT07}. ProbLog is a probabilistic language extending Prolog, where facts and clauses are annotated with (mutually independent) probabilities. 
Probabilistic logic programs based on distribution semantics can be viewed as a "programming language" generalization of Bayesian Networks~\cite{DBLP:conf/iclp/Sato95}.
Many PLP frameworks, e.g. ProbLog, ICL, PRISM and CP-Logic, do not support general \emph{normal} logic programs (cfr. Section~\ref{sec:semantics}). For instance ICL uses acyclic logic programs, while ProbLog's and CP-Logic's valid inputs are programs where each possible world corresponds to a two-valued unique well-founded model~\cite{DBLP:conf/pkdd/DriesKMRBVR15}. When general normal logic programs are concerned, well-founded semantics~\cite{DBLP:journals/jacm/GelderRS91} is a three-valued semantics, that is, logical atoms can be true, false or undefined. On the other hand, stable model semantics~\cite{DBLP:conf/iclp/GelfondL88} defines two-valued models, but a logic program can have more than one stable model. Both semantics have been considered to extend traditional PLP framework's semantics: \citet{DBLP:conf/ismvl/HadjichristodoulouW12} presents an extension of PRISM based on three-valued well-founded models. On the other hand, the application of stable model semantics in PLP is represented by probabilistic answer set programming (ASP) \cite{DBLP:journals/jair/CozmanM17,DBLP:journals/ijar/CozmanM20}.
With \smp{}, we consider stable model semantics in the context of the distribution semantics, that is, we extend a probability distribution over the probabilistic choices in the program to a probability distribution over stable models of the full program, whereas probabilistic ASP directly defines the latter; cf.~also Example~\ref{ex:global}. 

In this paper we will focus on computing exact probabilities as in \citet{DBLP:journals/tplp/BaralGR09} and~\citet{DBLP:conf/kr/LeeW16} rather than defining an interval of probability for each atom as in \emph{credal semantics}~\cite{DBLP:conf/ecai/Lukasiewicz98}.

We will consider the application of this semantics to probabilistic argumentation problems. This is a novel approach as previous work considers different semantics and reasoning techniques for probabilistic argument graphs. 
An \emph{argument graph}~\cite{DBLP:journals/ai/Dung95} is a pair $(A,R)$ where $A$ is a set of arguments and $R\subseteq A\times A$ is a binary (attack) relation over $A$. A \emph{probabilistic argument graph}~\cite{DBLP:conf/tafa/LiON11} is a tuple $(A,R,P_A,P_R)$ where: $(A,R)$ is an argument graph and $P_A$ and $P_R$ are functions: $P_A:A\rightarrow[0,1]$ and $P_R:A\times A\rightarrow[0,1]$.
The \emph{constellations}  approach~\cite{DBLP:journals/ijar/Hunter13} interprets the probabilities as uncertainty over the structure of the graph and models a probability distribution over subgraphs. 
The admissible arguments are defined by the classical extension-based semantics~\cite{DBLP:journals/ai/Dung95}. \citet{DBLP:journals/ijar/MantadelisB20} presents an encoding of a constellations approach in ProbLog.
The \emph{epistemic} approach interprets the probabilities as a direct measure of an agent's belief in the arguments. Previous work focused on reasoning about the properties of families of probability distributions that are consistent with the argument graph and additional constraints (\emph{epistemic graphs})~\cite{DBLP:journals/ai/HunterPT20}. Our approach combines both views into a novel reasoning framework. From the epistemic approach we adopt the view of probabilities as a measure of the degree of belief of an agent, but consider a single probability distribution in a Bayesian style rather than families of distributions. 
Such distribution, as in the constellation approach, is determined by the probability that an argument is true (accepted) in a possible world, defined by the distribution semantics. However, in order to determine whether an argument is true or not, we rely on the stable model semantics for logic programs rather than the classical argumentation semantics. 
Finally, \citet{DBLP:conf/ijcai/KidoO17} applies Bayesian reasoning techniques for statistically estimating the existence of an attack relation between arguments, but does not manipulate degrees of belief directly connected with acceptability.

\section{Probabilistic Semantics}\label{sec:semantics}

A \emph{normal} logic program $\mathcal{L}$ is a set of rules of the form $ h \leftarrow b_1,\dots,b_{n}$. We use the standard terminology about atoms, terms, predicates and literals. The \emph{head} $h$ is an atom and the \emph{body} $b_1,\dots,b_n$ is a logical conjunction of literals. We denote with $\lnot a$ the logical negation of an atom $a$, and with ${\sim}a$ the negation as failure of $a$. Rules with empty body are called facts.
We model uncertainty in logic programs by annotating facts with probabilities: the probabilities of facts are mutually independent and each \emph{probabilistic fact}, i.e. $p::f.$, corresponds to an atomic choice, that is, a choice between including $f$ in the program (with probability $p$) or discarding it ($1-p$). Probabilistic logic programs are queried for the likelihood of atoms, which is commonly defined by the \emph{distribution semantics}~\cite{DBLP:conf/iclp/Sato95}. A \emph{total choice} is a combination of atomic choices over all probabilistic facts, that is, a subset of the set of all facts $f$. We denote the set of all total choices of $\mathcal{L}$ with $\Omega(\mathcal{L})$. 
The probabilities of facts define a probability distribution over all ($2^n$ with $n$ probabilistic facts) total choices defining the possible non-probabilistic \emph{subprograms}, also called \emph{possible worlds}, obtained from $\mathcal{L}$ by including or discarding a fact according to the corresponding atomic choice. The probability of success of a query is the sum of the probability of each possible world where the query is true.

As the distribution semantics relies on a one-to-one mapping between total choices and models,
many PLP frameworks, e.g. ProbLog, PRISM and CP-Logic, restrict the class of valid inputs to programs where each total choice corresponds to a two-valued well founded model~\cite{DBLP:journals/jacm/GelderRS91}. As we will show, this limitation prevents the encoding of a variety of causal relations as far as argumentation is concerned. For this reason we focus on more general semantics to define models and probabilities based on total choices.
We consider \emph{stable model semantics}~\cite{DBLP:conf/iclp/GelfondL88}:
given a normal logic program $\mathcal{L}$ and a set $S$ of atoms interpreted as $true$, called \emph{candidate model}, stable model semantics is defined in terms of the \emph{reduct} of a program $\mathcal{L}$ w.r.t. $S$. The \emph{reduct} of $P$ w.r.t. $S$, $\mathcal{L}^S$, is defined as follows:
1) for all $r\in \mathcal{L}$ remove $r$ from $\mathcal{L}$ if $a\in S$ is negated in the body of $r$;
2) for all $r\in \mathcal{L}$ remove from the body of $r$ all $\lnot a, a\notin S$.
If $S$ is a minimal model for $\mathcal{L}^S$ then $S$ is a \emph{stable model} (\emph{answer set}) for $\mathcal{L}$. 

Using stable models allows us to relax the standard PLP assumption of exactly one model per total choice.
\begin{definition}
A valid \smp{} program is a probabilistic normal logic program where each total choice leads to \emph{at least} one (two-valued) stable model.
\end{definition} 
We note that this definition excludes the case where a total choice has \emph{no} stable model, i.e., results in an inconsistent program, and thus a loss of probability mass. Our focus here is on handling non-probabilistic choices introduced by the logic component, as illustrated in the following example.

\begin{example}\label{ex:cycle}
Consider the following normal logic program:
$0.5::a. \;\; 0.5::b.\quad c\leftarrow a.\quad d\leftarrow b.\quad c\leftarrow {\sim}d. \quad d\leftarrow {\sim} c.$
There are four possible worlds: $\omega_1=\{a,b\}$, $\omega_2=\{a\}$, $\omega_3=\{b\}$, $\omega_4=\{\}$, with $P(\omega_i)=0.25$. While the first three correspond to one stable model, i.e. $M(\omega_1)=\{\{a,b,c,d\}\}$, $M(\omega_2)=\{\{a,c\}\}$, $M(\omega_3)=\{\{b,d\}\}$, $\omega_4$ has two stable models: $M(\omega_4)=\{\{c\},\{d\}\}$.
\end{example}

We follow the principle of maximum entropy and choose to uniformly distribute the probability of a total choice across all the corresponding stable models. Consider for instance Example~\ref{ex:cycle}: while the probabilistic choices characterize the four possible worlds, the logic encoded in the rules prescribes in $\omega_4$ a choice between $c$ and $d$. Since this choice is not related to beliefs but rather logical consistency, we assume that all stable models are equally probable for a given total choice. More formally, each total choice $\omega$ corresponds to: (1) a probability $P(\omega)$ as defined by distributions semantics and (2) a set $M(\omega)$ of stable models: 
\begin{definition}
Given a probabilistic normal logic program $\mathcal{L}$, the probability $ P(\omega)$ of a total choice $\omega$ is:
\[P(\omega)=\prod_{(f:p)\in\mathcal{L}, f\in\omega} p\cdot\prod_{(f:p)\in\mathcal{L}, f\not\in\omega} 1-p\]
\end{definition}
that is, the product of the probabilities of the facts for being included/excluded from $\omega$.

\begin{definition}
Given a probabilistic normal logic program $\mathcal{L}$ and a corresponding total choice $\omega$, $M(\omega)$ is the set of stable models of the subprogram induced by $\omega$ on $\mathcal{L}$.
\end{definition}

The probability of a stable model $S_{\omega}\in M(\omega)$ is the probability of the corresponding total choice $\omega$ normalized w.r.t. the number of the stable models for that possible world:
\begin{definition}
Given a probabilistic normal logic program $\mathcal{L}$ and a total choice $\omega$, $\forall S_{\omega}\in M(\omega)$:
$\hat{P}(S_{\omega}) = \frac{P(\omega)}{|M(\omega)|}$
\end{definition}

The probability of an atom $a$ is the sum of the probabilities of the models where $a$ is true:
\begin{definition}
Given a probabilistic normal logic program $\mathcal{L}$, the probability $\mathbb{P}(a)$ of success of querying $a$ in $\mathcal{L}$ is:
\[\mathbb{P}(a)=\sum_{a\in S_{\omega},S_{\omega}\in M(\omega),\omega\in\Omega(\mathcal{L})}\hat{P}(S_{\omega})\]
\end{definition}

Our approach generalizes traditional PLP frameworks: when a program defines total choices corresponding to exactly one two-valued well-founded model, then the semantics agree on both models and probability.
As for the model, if a normal logic program has a total two-valued well-founded model, then the model is the unique stable model~\cite{DBLP:journals/jacm/GelderRS91}. Thus, in this case $|M(\omega)|=1$ for each total choice $\omega$, from which it follows that $\hat{P}(S_{\omega})=P(\omega)$ for the single $S_{\omega}\in M(\omega)$ ($M(\omega) = \{S_{\omega}\}$). This means that the probability of the model is the probability of the corresponding total choice as in (probabilistic) two-valued well founded semantics.
\begin{example}
We complete Example~\ref{ex:cycle}: models $\{a,b,c,d\}$, $\{a,c\}$, $\{b,d\}$ are the unique model for, respectively, $\omega_1,\omega_2,\omega_3$ hence $\hat{P}(\{a,b,c,d\})=\hat{P}(\{a,c\})=\hat{P}(\{b,d\})=P(\omega_1)=P(\omega_2)=P(\omega_3)=0.25$. The likelihood of $\omega_4$ is uniformly distributed over the two stable models $\{c\}$ and $\{d\}$, whose probability is thus $\frac{P(\omega_4)}{|M(\omega_4)|}=\frac{0.25}{2}=0.125$. Therefore
$\mathbb{P}(a)=\mathbb{P}(b)=0.5$, $\mathbb{P}(c)=\mathbb{P}(d)=0.625$.
\end{example}

The key difference between \smp{} and probabilistic ASP frameworks is that our approach is based on the distribution semantics, that is, we consistently extend a probability distribution over total choices to a probability distribution over models of the logic program, whereas probabilistic ASP languages such as P-Log~\cite{DBLP:journals/tplp/BaralGR09} and LP\textsuperscript{MLN}~\citet{DBLP:conf/kr/LeeW16}  use the probability labels to directly define a (globally normalized) probability distribution over the models of a (derived) program.
Consider the following example:
\begin{example}\label{ex:global}
In Example~\ref{ex:cycle}, if we associate to each stable model $S_{\omega}$ the corresponding $P(\omega)$, the sum of the probabilities of the stable models is $5\cdot0.25=1.25$ and a normalization w.r.t. all answer sets leads to a probability $\hat{P}'(S_{\omega})=0.25/1.25=0.2$ for all $S_{\omega}$. Then $\mathbb{P}(a)=\hat{P}'(S_{\omega_1})+\hat{P}'(S_{\omega_2})=0.4$. This means that the marginal probability of $a$ in the joint distribution is lower than the prior 0.5 despite no epistemic influence is defined on $a$. In our approach, the  marginal probability of that fact under the final distribution equals the fact's label, and is thus directly interpretable.
\end{example}

\section{Inference and learning}\label{sec:implementation}
In this section we present the algorithms implemented in \smp{} to perform inference and learning tasks. We reduce the probabilistic inference task to a \emph{weighted model counting problem} (WMC)~\cite{DBLP:journals/aicom/CadoliD97}. The task of model counting is \#P-complete in general, therefore we follow the approach of~\citet{DBLP:conf/uai/FierensBTGR11} of transforming the logic program into a representation where this task becomes tractable. In this setting, the original grounding procedure is no longer applicable, since it uses the presence of cycles through negation as a sufficient condition to reject programs for which the two-valued well-founded models semantics is not well-defined. We describe the novel normalization algorithm and show that under this semantics ProbLog2's expectation-maximization algorithm  (EM learning) for Bayesian learning over a set of evidence~\cite{DBLP:conf/pkdd/GutmannTR11} is correct.
\subsection{Inference}
The inference task is defined as follows: 
\begin{definition}
    The Inference task: \textbf{Given}
\begin{itemize}
    \item[-] A ProbLog program $\mathcal{L}$: let $G$ be the set of all ground (probabilistic and derived) atoms of $\mathcal{L}$.
    \item[-] A set $E\subseteq G$ of observed atoms (evidence), along with a vector $e$ of corresponding observed truth values ($E = e$).
    \item[-] A set $Q\subseteq G$ of atoms of interest (queries).
\end{itemize}
\textbf{Find} 
the marginal distribution of every query atom given the evidence,
i.e. computing $P(q\,|\,E = e)$ for each $q \in Q$.
\end{definition}
\paragraph{Compilation.} In \smp{} the reduction to WMC is based on the pipeline described in~\citet{DBLP:conf/uai/FierensBTGR11}, with the adaptations proposed in~\citet{DBLP:conf/aaai/AzizCMS15}, where the CNF conversion of $\mathcal{L}$ is replaced with an encoding of the rules, variables and the corresponding strongly connected components as the input for the knowledge compiler. $\mathcal{L}$ is in fact transformed by means of knowledge compilation into a \emph{smooth d-DNNF} formula (deterministic Decomposable Negation Normal Form)~\cite{DBLP:conf/ecai/Darwiche04}.
\begin{definition}
A \emph{NNF} is a rooted directed acyclic graph in which each leaf node is labeled with a literal and each internal node is labeled with a disjunction or conjunction. A smooth \emph{d-DNNF} is an NNF with the following properties:
\begin{itemize}
    \item Deterministic: for all disjunctive nodes the children represent formulas pairwise inconsistent.
    \item Decomposable: the subtrees rooted in two children of a conjunction node do not have atoms in common.
    \item Smooth: all children of a disjunction node use the same set of atoms.
\end{itemize} 
\end{definition}
On d-DNNFs the task of model counting becomes tractable.
The d-DNNF is further transformed into an equivalent arithmetic circuit, by replacing conjunctions and disjunctions respectively with multiplication and summation nodes, and by replacing leaves with the weight of the corresponding literals. The arithmetic circuit thus allows us to efficiently perform the task of WMC.
In this pipeline, we replace the grounding step with a bottom-up grounding procedure to handle normal logic programs, and use the absence of cycles through negations as a condition to reduce to ProbLog's evaluation algorithm, where \smp's semantics is equivalent as discussed in Section~\ref{sec:semantics}. If this is not the case, we evaluate the output of the pipeline as follows.
\paragraph{Enumeration.} 
The new semantics requires us to know for each model the corresponding normalization constant. We derive from the d-DNNF an equivalent circuit where each leaf for a literal $l$ is replaced by a list of (partial) models $[\{l\}]$, disjunctions are replaced by the union of the children and conjunctions correspond to the cartesian product of the children. 
Traversing bottom-up such circuit returns the list of models, from which we build a map from total choices to the corresponding (number of) models. In fact, for each model the truth value of the literals corresponding to probabilistic facts defines the corresponding total choice. 
This is a computationally expensive step since it entails enumerating all possible models.
\paragraph{Evaluation.} To evaluate $P(q\,|\,E=e)$, the weights of the leaves in the arithmetic circuit are instantiated according to the query and the given evidence.  The weight of the root is then the sum of the weights of the models where $q$ is true and evidence is satisfied. Since WMC can be performed efficiently on the circuit, we compute first the unnormalized weight of the root  $w^*$, and then correct it. We know that for each total choice $\omega$ s.t. $n=|M(\omega)|>1$ and $\omega\models E=e$, we are overcounting the weight of its models by $w_{M(\omega)}=\frac{P(\omega)\cdot(n-1)}{n}$, hence we remove $w_{M(\omega)}$ from $w^*$ for each $S_{\omega}\in M(\omega)$ such that $S_{\omega}\models q\land E=e$ (Algorithm~\ref{algo:evaluate}). This correction increases the complexity of the evaluation step, as we need to iterate over the total choices with multiple models to retrieve and apply the different normalization constants. Normalizing weights while traversing the circuit is possible but less efficient. 
In fact, in the d-DNNF different total choices can correspond to the same node, e.g. when conjoining a literal and a disjunction node, therefore the weight of such node has to be normalized with (potentially) different constants for its children. At the same time normalization cannot be applied on the children level, because they might be a subtree for different total choices. Therefore, their weights cannot be aggregated in a single value until a total choice is defined, which is a similar process to the enumeration step on the d-DNNF. In Section~\ref{sec:experiments} we show that exploiting the efficiency of the (unnormalized) WMC task results in the evaluation step being significantly faster than the enumeration step.

\begin{algorithm}
    \caption{Evaluation step: $P(q | E=e)$}\label{alg:cap}
    \begin{algorithmic}
    \Ensure $P(q\,|\,E=e)$
    \State $w^* \gets unnormalized(root(\mathcal{C}))$
        \For{$\omega$ \textbf{in} $\Omega(\mathcal{L})$ \textbf{s.t.} $|M(\omega)|>1,\omega\models E=e$ }
            \For{$S_{\omega}\in M(\omega)$ \textbf{s.t.} $ S_{\omega}\models q\land E=e$}
                    \State $w^* = w^*-\frac{P(\omega)\cdot(n-1)}{n}$
            \EndFor
        \EndFor
    \State \Return $w^*$
    \end{algorithmic}
    \caption{Evaluation step schema}
    \label{algo:evaluate}
\end{algorithm}

\subsection{Learning by Expectation Maximization} \label{sec:learning}
In the learning from interpretation setting, given a ProbLog program $\mathcal{L}$ where some probabilities are unknown (parameters), and a set of interpretations for $\mathcal{L}$, the goal is to estimate the value of the parameters such that the predicted values maximize the likelihood of the given interpretations:
\begin{definition}
    Max-Likelihood Parameter Estimation: \textbf{Given}
\begin{itemize}
    \item[-] A ProbLog program $\mathcal{L}\mathbf{(p)} = \mathcal{F} {\cup}\mathcal{BK}$ where $\mathcal{F}$ contains probabilistic facts and $\mathcal{BK}$ contains background knowledge. $\mathbf{p} = \langle p_1, ..., p_N\rangle$ is a set of unknown parameters attached to probabilistic facts.
    \item[-] A set $\mathcal{I}$ of (partial) interpretations $\{I_1, ..., I_M\}$ as training examples.
\end{itemize}
\textbf{Find} 
the maximum likelihood probabilities $\widehat{\mathbf{p}} = \langle \widehat{p_1}, ..., \widehat{p_N}\rangle$ for all interpretations in $\mathcal{I}$. Formally,
\[
    \widehat{\mathbf{p}} = \arg \max_\mathbf{p} P(\mathcal{I|L}\mathbf{(p)}) = \arg\max_\mathbf{p} \prod^M_{m=1} P_s(I_m|\mathcal{L}\mathbf{(p)})
\]
\end{definition}
Learning from interpretations of parameters in a ProbLog program is implemented in a likelihood maximization setting such that the parameters are iteratively updated. 

In total observability each interpretation $I_j\in \mathcal{I}$ observes the truth value of each atom and probabilistic fact of $\mathcal{L}$. This case reduces to counting the number of true occurrences of each of probabilistic fact in the interpretations $\mathcal{I}$~\cite{DBLP:conf/pkdd/GutmannTR11}. This is correct for \smp{} because the observation of a probabilistic fact is independent of the rest of the program for each interpretation.


In the partially observable case, where each interpretation $I_j$ observes the truth value of a subset of $\mathcal{L}$, the parameter update iteration relies on probabilistic inference to update the likelihood of a fact given an interpretation $I_m$. 
At each iteration $k$ the parameters from the previous iteration $k-1$ are used in $\mathcal{L}$ to compute the conditional expectation of the parameter given the interpretations, until convergence. 
This means that the normalization w.r.t. multiple stable models is incorporated in the conditional probabilities computed by the inference task, therefore also the parameter estimate is normalized w.r.t. the different stable models that may correspond to a given partial observation.





\section{Modelling Probabilistic Argumentation}\label{sec:args}
In this section we show how PLP modelling techniques can be applied to encode causal relations between beliefs in arguments. We show how causal relations between arguments often lead to models that require the generalized semantics presented in the previous section. The flexibility of PLP not only captures basic probabilistic argument graph, but also many other features from the argumentation literature:
\begin{itemize}
    \item Support relations.
    \item Quantitative evaluations of attacks/supports.
    \item Attack (and supports) from sets of arguments.
    \item Distinctions between proponents of arguments.
\end{itemize}
which we also combine with features from PLP  like first-order knowledge and querying conditional probabilities. 
Finally, we show an intuitive method to model belief in arguments and their relations, by means of PLP and ProbLog's syntactical features, to provide an implementation of our approach. 
Let us consider Example~\ref{ex:microtext} to describe our modelling approach. We model arguments ($a_i$), attacks (relation $R^-$) $\{(a_1,a_6),(a_4,a_1)\}$,$(a_1,a_2),(a_2,a_1)\}$, supports (relation $R^+$) $\{(a_5,a_4), (a_3,a_1)\}$ and distinguish between arguments coming from a proponent ($\{a_2,a_4,a_5,a_6\}$) and those introduced by an opponent ($\{a_1,a_3\}$).
We also define the following probability functions to derive an example of a probabilistic argumentation problem: $P_A(a_1)=0.4$, $P_A(a_2)=0.8$, $P_A(a_3)=0.3$, $P_A(a_4)=0.7$, $P_A(a_5)=0.6$, $P_A(a_6)=0.7$, $P_{R^-}(a_1,a_6)=0.6$, $P_{R^-}(a_2,a_1)=0.8$, $P_{R^-}(a_1,a_2)=0.7$, $P_{R^-}(a_4,a_1)=0.3$, $P_{R^+}(a_5,a_4)=0.6$,  $P_{R^+}(a_3,a_1)=0.5$. Note that we already consider probabilistic argument graphs where a (probabilistic) support relation is also included.

We define the semantics of the probabilities in an argument graph by mapping each element of the graph to a fact, rule or probability in a PLP model. 
We model a probabilistic argumentation graph (with supports) by mapping:
\begin{enumerate}
    \item $P_A(a)=p$ to the fact $p::base\_arg(a)$ and the rule $arg(a)\leftarrow base\_arg(a).$
    \item  attacks $P_{R^-}(a,b)=p$ to rules  $p::\lnot arg(b) \leftarrow arg(a).$
    \item  supports $P_{R^+}(a,b)=p$ to rules $p:: arg(b) \leftarrow arg(a).$
\end{enumerate}
Arguments thus map to predicates of the form $arg(a)$, which we will query for their truth value. The semantics of $P_A$ is given by the probabilistic facts of the form $p::base\_arg(a)$: this is a prior belief from which the truth of the corresponding argument can be inferred. It is independent of the other epistemic information, therefore represents an unconditioned bias towards $arg(a)$.
The edges of an argument graph model the influence between the arguments: we encode them as \emph{causal relations} between the acceptance of two arguments. That is, a support $arg(b) \leftarrow arg(a).$ reads as ``accepting $a$ causes believing in (accepting) $b$''. Viceversa, $\lnot arg(b) \leftarrow arg(a).$ reads as ``accepting $a$ causes not believing in (rejecting) $b$''. The semantics of $P_{R^+}$ and $P_{R^-}$ is given by the mapping of the probability of the relation as the probability of the rule. 
For example a probability $P_{R^+}(a_5,a_4)=0.6$ is mapped to $0.6::arg(a_4) \leftarrow arg(a_5)$, which describes how believing in $a_5$ causes a (fine-grained) increase in the belief in $a_4$.
Similarly, attacks are encoded as supports for counterarguments, resulting in an inhibition of the belief in the original argument, e.g. $0.3::\lnot arg(a_1) \leftarrow arg(a_4).$
This mapping relies on three features of PLP (and ProbLog): first, rule heads can be annotated as syntactic sugar: $p::h\leftarrow b_1,\dots,b_n.$ is equivalent to a new fact $p::f.$ plus $h\leftarrow f, b_1,\dots,b_n.$ This gives a quantitative evaluation of how likely believing in the body causes believing the head as well.
Second, attacks define an \emph{inhibition effect}~\cite{DBLP:conf/pgm/MeertV14} on a belief by means of a language feature for probabilistic logic frameworks called \emph{negation in the head}~\cite{DBLP:journals/corr/Vennekens13}. Negation in the head gives an interpretation of the negation of heads from Answer Set Programming in the context of epistemic reasoning and logic theories defining causal mechanisms. We describe such interpretation by means of ProbLog's syntax: each rule $\lnot h\leftarrow b_1,\dots,b_n.$ is interpreted by replacing all heads $h$ in the program with a new atom $h_{pos}$ and all heads $\lnot h$ with a new atom $h_{neg}$, and the rule $h\leftarrow h_{pos},{\sim}h_{neg}.$ is added. This last rule thus defines that $h$ can be inferred from any of the causes of $h$ (making $h_{pos}$ true) only if there is no cause for believing in $h_{neg}$, the opposite of $h$. Finally, the probability of an atom which appears as the head of multiple rules is computed by means of the typical \emph{noisy-or} effect~\cite{DBLP:conf/pgm/MeertV14}. 
\begin{example}\label{ex:rewrite}
We show the rewriting for $P_A(a_1)=0.4$, $P_A(a_2)=0.8$ and $P_{R^-}(a_1,a_2)=0.7$:\\
$
\begin{array}{ll}
    \multicolumn{2}{l}{0.4::base\_arg(a_1). \quad 0.8::base\_arg(a_2).\quad 0.7::f.}\\ 
    arg(a_2)\leftarrow base\_arg(a_2). & arg(a_1)\leftarrow base\_arg(a_1).\\
    \multicolumn{2}{l}{arg(a_2) \leftarrow arg_{pos}(a_2), {\sim}arg_{neg}(a_2).}\\
    arg_{neg}(a_2) \leftarrow f, arg(a_1). &  
\end{array}   
$
\end{example}

These features combined provide the semantics for an argument graph where the belief of an argument $a$ is defined by its marginal probability in the modelled joint distribution. This belief can be queried in a PLP style, which thus reflects the combined interaction of a bias towards $a$ with the belief in other arguments and their relation with $a$.

\begin{example}\label{ex:enccycle}
We model our running example as:
\[
\begin{array}{ll}
    \multicolumn{2}{l}{0.4::base\_arg(a_1). \quad 0.8::base\_arg(a_2).}\\
    \multicolumn{2}{l}{0.3::base\_arg(a_3).\quad 0.7::base\_arg(a_4). }\\
    \multicolumn{2}{l}{0.6::base\_arg(a_5). \quad 0.7::base\_arg(a_6).}\\
    arg(A)\leftarrow base\_arg(A). & \\
    0.6::\lnot arg(a_6) \leftarrow arg(a_1). & 0.6::arg(a_4) \leftarrow arg(a_5).\\
    0.3::\lnot arg(a_1) \leftarrow arg(a_4). & 0.5::arg(a_1) \leftarrow arg(a_3). \\
    0.8::\lnot arg(a_1) \leftarrow arg(a_2). & 0.7::\lnot arg(a_2) \leftarrow arg(a_1).
\end{array}   
\]
Resulting in the distribution (beliefs): $\mathbb{P}(arg(a_1))=0.21$, $\mathbb{P}(arg(a_2))=0.69$, $\mathbb{P}(arg(a_3))=0.3$, $\mathbb{P}(arg(a_4))=0.81$, $\mathbb{P}(arg(a_5))=0.6$, $\mathbb{P}(arg(a_6))=0.61$.
\end{example}

Example~\ref{ex:enccycle} shows how arguments attacking each other results in a cyclic relation where negation is involved, similar to the one presented in Example~\ref{ex:cycle}. Therefore, such interpretation of an argument graph would not be possible with traditional PLP semantics.

\begin{example}
    In Example~\ref{ex:enccycle} rules 
    $0.8::\lnot arg(a_1) \leftarrow arg(a_2).$ and $0.7::\lnot arg(a_2) \leftarrow arg(a_1).$ correspond to the rewriting of Example~\ref{ex:rewrite} plus the following statements:
    \[
    \begin{array}{l}
        0.8::g.\quad arg_{neg}(a_1) \leftarrow g, arg(a_2). \\ 
        arg(a_1) \leftarrow arg_{pos}(a_1), {\sim}arg_{neg}(a_1).\\
    \end{array}   
    \]
    The negation in the head thus leads to a cyclic dependency $arg(a_1)\rightarrow^- arg_{neg}(a_1)\rightarrow arg(a_2)\rightarrow^- arg_{neg}(a_2) \rightarrow arg(a_1)$ which results in possible worlds with multiple stable models (thus invalid for ProbLog's semantics). 
\end{example}


The logic program not only can be queried for the marginal probabilities of the arguments, but also for conditional probabilities by specifying \emph{evidence}, that is, a set of facts for which the truth value is determined or assumed. By means of evidence the joint probability distribution can answer queries of the form $\mathbb{P}(arg(a_i)|arg(a_j))$ which thus answer the question ``What  is the belief in $a_i$  if 
 $a_j$ is known to hold?''. This allows an update of the beliefs in arguments in the light of new information about their acceptability. 
\begin{example}
By querying the model from Example~\ref{ex:enccycle} with the addition of the statement $evidence(arg(a_1), true).$ we infer the following conditional probabilities: $\mathbb{P}(arg(a_1)|arg(a_1))=1$, $\mathbb{P}(arg(a_2)|arg(a_1))=0.12$, $\mathbb{P}(arg(a_3)|arg(a_1))=0.43$, $\mathbb{P}(arg(a_4)|arg(a_1))=0.75$, $\mathbb{P}(arg(a_5)|arg(a_1))=0.58$, $\mathbb{P}(arg(a_6)|arg(a_1))=0.28$.
\end{example}
Our framework includes two of the most relevant extensions of the basic argument graph, namely  \emph{bipolar} argumentation~\cite{DBLP:conf/nmr/AmgoudCL04}, which introduces the support relation, and 
\emph{weighted} argumentation~\cite{DBLP:journals/ai/DunneHMPW11,DBLP:conf/kr/Coste-MarquisKMO12}, which introduces a fine-grained quantitative evaluation of the relations between arguments.
Moreover, we can expand our framework further to \emph{argument systems} ~\cite{DBLP:conf/argmas/NielsenP06}, which generalize the attack between two arguments to an attack relation from a set of arguments towards a single argument (set-attacks). The belief in an argument is thus inhibited only if all the attackers of the set are believed, which we model by conjoining attackers in the body of rules, e.g. $\lnot arg(a_1) \leftarrow arg(a_4), arg(a_5).$ Clearly, also in this case we can easily combine different argumentative features with each other, for example gradual set attacks, e.g. $0.6::\lnot arg(a_1)\leftarrow arg(a_4),arg(a_5).$ or (gradual) set supports, e.g. $0.3::arg(a_3) \leftarrow arg(a_4), arg(a_5).$ 
Thanks to the generality of PLP we can also model less standard scenarios. For example bias towards arguments can be defined in terms of trust in the agent proposing the argument, with rules $0.4::prop. \;\;0.6::opp.$ and: 
\[
\begin{array}{c}
     \;\; arg(a_1)\leftarrow opp.\;  arg(a_2)\leftarrow prop. \; \arg(a_3)\leftarrow opp. \\
     arg(a_4)\leftarrow prop. \; arg(a_5)\leftarrow prop. \; arg(a_6)\leftarrow prop.
\end{array}   
\]
Moreover, PLP systems are designed to express first-order knowledge, allowing a compact encoding of high-level reasoning, as we already showed in Example~\ref{ex:enccycle} with the rule $arg(A)\leftarrow base\_arg(A).$ For instance, we can model that the more agents propose the same argument, the higher its bias is. Assume we express proponents and trust as $p_1::prop(1).,\dots,p_n::prop(n).$ and $proposes(i,a)$ denotes that argument $a$ is backed by proponent $i$, then a rule $base\_arg(A) \leftarrow proposes(P,A), prop(P).$ compactly defines a contribution to the bias of each argument, denoted by each measure of trust $p_i$, from all its proponents.

\section{Experiments}\label{sec:experiments}

The goal of our experiments is to establish the feasibility of our approach, and to answer the following questions: (Q1) what is the performance impact of the more general semantics on the inference task? (Q2) Given a set of observations of accepted arguments generated from a known ProbLog program, can we learn the degrees of belief of the agent that modelled the original program?

\paragraph{Q1. Inference.} 
We consider a variation of a typical PLP example where a set of people has a certain probability of having asthma or being stressed, and stress leads with some probability to smoking:
\[
\begin{array}{l}
    0.1::asthma(X)\leftarrow \mathit{person}(X).\\
    0.3::stress(X)\leftarrow \mathit{person}(X).\\
    0.4::smokes(X)\leftarrow stress(X).
\end{array}   
\]
People are related by an influence relation: if a person smokes and influences to some extent another one, then the other person will smoke, and if someone smokes there is a probability to suffer from asthma. If someone suffers from asthma then the person does not smoke (an example of a cycle with negation).
\[
\begin{array}{l}
    smokes(X) \leftarrow \mathit{influences}(Y,X), smokes(Y).\\
    0.4::asthma(X) \leftarrow smokes(X). \\
    \lnot smokes(X) \leftarrow asthma(X).
\end{array}   
\]
We consider examples with an increasing number of people and relationships: $t_1$ = $\{\mathit{person}(1).\;\mathit{person}(2).\;0.3::\mathit{influences}(1,2).\;0.6::\mathit{influences}(2,1).\}$, $t_2=t_1+\{\mathit{person}(3)\}$, $t_3=t_2+\{\mathit{person}(4)$\}, $t_4=t_3+\{0.2::\mathit{influences}(2,3).\}$, $t_5=t_4 + \{0.7::\mathit{influences}(3,4).\}$, $t_6=t_5+\{0.9::\mathit{influences}(4,1).\}$. Figure (\subref{fig:inference}) shows the running time of \smp{} on the different benchmarks, where the enumeration step dominates the running time of \smp{}, while the evaluation step adds further computational time, but not to the same extent.

\paragraph{Q2. Learning.}
We answer question 2 by considering a dataset of 283 argument graphs~\cite{DBLP:conf/lrec/StedeAPAP16} and by deriving from each annotation a bipolar argument graph. We attach to arguments and relations random probabilities to reflect an agent's belief in each, as we did in Example~\ref{ex:enccycle}. The graphs contain from 4 to 11 nodes (average $5.3$) with an average of $1.4$ attacks and $3$ supports per graph, which translates in about 10 learnable parameters on average  for each example.
We learn such parameters from a set of increasing size of observations of accepted arguments, sampled from the original program, with an upper bound of 100 iterations. Note that we are in the case of partial observability, as the parameters are attached to the predicates $base\_arg$, $arg_{pos}$ and ${arg_{neg}}$, but we observe only the outcome in the form of the predicates $arg$.
We evaluate the quality of the learned programs with the mean absolute error (MAE), summarized in  Figure (\subref{fig:learning}). Since the MAE decreases when more examples are given, we can answer positively to our question about learnability of beliefs in argument graphs with \smp{}.
\begin{figure}[t]
    \centering
        \begin{subfigure}[b]{\columnwidth} 
        \begin{tikzpicture}
            \pgfplotsset{%
                width=\columnwidth,
                height=0.7\textwidth
            }
            \begin{axis}[
                ybar stacked,
                bar width=15pt,
                ymajorgrids,
                grid style=dashed,
                enlargelimits=0.15,
                legend style={at={(0.45,0.85)},
                anchor=north,legend columns=-1},
                ylabel={time (s)},
                symbolic x coords={$t_1$,$t_2$,$t_3$,$t_4$,$t_5$,$t_6$},
                xtick=data,
                x tick label style={rotate=45,anchor=east},
                ytick={0,10,20,30,40}
                ]
            \addplot+[ybar] plot coordinates {($t_1$,0.006) ($t_2$,0.006) 
            ($t_3$,0.007) ($t_4$,0.007) ($t_5$,0.012) ($t_6$,0.019) };
            \addplot+[ybar] plot coordinates {($t_1$,0.008) ($t_2$,0.116) 
                ($t_3$,2.158) ($t_4$,5.337) ($t_5$,14.449) ($t_6$,32.541) };
                \addplot+[ybar] plot coordinates {($t_1$,0.110) ($t_2$,0.603) 
                ($t_3$,1.546) ($t_4$,2.158) ($t_5$,2.982) ($t_6$,7.460) };
            \legend{\strut compile, \strut enumerate, \strut evaluate}
            \end{axis}
        \end{tikzpicture}
        \begin{tabular}{|c|c|c|c|c|c|c|}
            \hline
            Benchmark & $t_1$ & $t_2$ & $t_3$ & $t_4$ & $t_5$ & $t_6$ \\\hline
            \# prob. facts & 10 & 14 & 18 & 19 & 20 & 21 \\\hline
            \# nodes circuit & 156 & 192 & 228 & 462 & 750 & 1465 \\\hline
        \end{tabular}  
        \caption{Inference time on benchmarks.}
        \label{fig:inference}
        \begin{subfigure}[b]{\columnwidth}
    \end{subfigure}
    
    \begin{tikzpicture}
        \pgfplotsset{%
            width=\columnwidth,
            height=0.52\textwidth
        }
        \begin{axis}[
            y tick label style={
                    /pgf/number format/.cd,
                    fixed,
                    fixed zerofill,
                    precision=2,
                    /tikz/.cd
            },
            xlabel={Number of samples},
            ylabel={MAE},
            xmin=0, xmax=350,
            ymin=0, ymax=0.15,
            xtick={0,50,100,150,200,250,300},
            ytick={0,0.05,0.1,0.15},
            legend pos=north west,
            ymajorgrids=true,
            grid style=dashed,
        ]
        
        \addplot[
            color=blue,
            mark=square,
            ]
            coordinates {
            (50,0.123)(100,0.086)(150,0.068)(200,0.062)(250,0.055)(300,0.052)
            };
        
        \end{axis}
    \end{tikzpicture}  
    \caption{Mean absolute error on learned parameters.}
    \label{fig:learning}
\end{subfigure}

\end{figure}         

\section{Conclusion}
Approaching probabilistic argumentation from a probabilistic logic programming perspective stresses the limiting assumptions of PLP frameworks when (probabilistic) normal logic programs are concerned. For this reason in this paper we proposed a new PLP system, \smp{}, where we implement a combination of the classical distribution semantics for probabilistic logic programs with stable model semantics. \smp{} generalizes previous work on distributions semantics supporting inference and parameter learning tasks for a wider class of (probabilistic) logic programs.

At the same time, we propose a mapping from probabilistic argument graphs to probabilistic logic programs, which provides a novel semantics for epistemic argument graphs. With our methodology, it is possible to apply traditional Bayesian reasoning techniques to probabilistic argument graphs, such as conditioning over evidence and parameter learning.
This results in a modular, expressive, extensible framework reflecting the dynamic nature of beliefs in an argumentation process.


\section*{Acknowledgements}
This work was supported by the FWO project N. G066818N.

\bibliography{biblio}

\end{document}